\documentclass{llncs}
\usepackage{graphicx}
\usepackage{amsmath}
\usepackage{color}

\begin{document}

\pagestyle{empty}

\mainmatter

\title{Robust Lane Tracking with Multi-mode Observation Model and Particle Filtering}

\titlerunning{Lecture Notes in Computer Science}

%

\author{Jiawei Huang\inst{1} \and Zhaowen Wang\inst{2}}
\institute{Honda Research Institute, USA\\
\email{jhuang@hra.com}\\
\and
Adobe Research\\
\email{zhawang@adobe.com}}



\maketitle

\begin{abstract}
Automatic lane tracking involves estimating the underlying signal from a sequence of noisy signal observations. Many models and methods have been proposed for lane tracking, and dynamic targets tracking in general. The Kalman Filter is a widely used method that works well on linear Gaussian models. But this paper shows that Kalman Filter is not suitable for lane tracking, because its Gaussian observation model cannot faithfully represent the procured observations. We propose using a Particle Filter on top of a novel multiple mode observation model. Experiments show that our method produces superior performance to a conventional Kalman Filter.
\end{abstract}

\section{Introduction}
\label{sect:intro}

Intelligent vehicles have important applications in autonomous steering and
driver status monitoring.
Such vehicles can greatly improve vehicle control and road safety.
Increasing computational power and universal access to
low cost cameras have stimulated
research interest in intelligent visual systems.
Many prototype systems implemented in previous years such as AURORA system \cite{AURORA97} designed by the
Robotics Institute of CMU,
and the DARPA Grand Challenge winner Stanley \cite{Stanley06}, developed by
Stanford's Artificial Intelligence Lab are well tested for specific tasks, but lack commercial viability because they fail to address issues such as unpredictable road condition, unstable vehicle speed,
varying weather and lighting, presence of other vehicles, etc.

The primary objective of an intelligent vehicle visual system is to identify the land mark and locate the lane position.
Images are captured using a mounted video camera, processed using edge detection and analyzed
to determine lane information.
Given a specific lane model, a detector searches for the best model parameter that matches
the edge points found in a single frame.
The simplest model - straight line - can be easily detected through Hough transform \cite{LineLane05}.
More complex algorithms, such as B-Snake \cite{BSnakeLane00} and parabola \cite{ParaLane06},
more accurately simulates the lane which is detected by optimizing likelihood functions
with MAP estimation.

The performance of detection based methods is heavily affected by image noise.
To improve robustness, lane-tracking algorithms are used to consider the temporal transition of the lane state.
Kalman Filter (KF) is used extensively in lane tracking applications
\cite{LineLane05,LearnKF06,SysKF01}
for its optimality in linear Gaussian processes.
At each time epoch, a detection result is fed into KF as a single sample
of the observation distribution, and the predicted distribution is updated to
get the optimal posterior estimation in terms of Minimum Mean Square Error (MMSE).

However, the Gaussian observation model in KF is often inadequate in representing
detection results on complex images.
Hence more informative observation models with unrestricted forms are proposed.
In \cite{ParaLane06,TrackingPF07}, the likelihood of a candidate lane state is
proportional to the edge strength within the region covered by the lane.
In \cite{MultiCuePF03}, multiple cues including color and gradient are fused together
for a more robust observation in cluttered environment.
As the resulting observations are no longer linear and Gaussian, Particle Filter (PF)
has been used in all the works above \cite{ParaLane06,TrackingPF07,MultiCuePF03}.
PF is able to approximate target distribution with Monte Carlo sampling, thus allowing
arbitrary form of observation distribution.

In this paper, we further investigate the problem of lane tracking with PF.
A novel multiple mode observation model is proposed by jointly considering several
tentative detection results.
This approach effectively preserves the information found in the detection step.
Moreover, it is computationally more economic than the observation models in
\cite{ParaLane06,TrackingPF07,MultiCuePF03},
which must be calculated separately for each candidate state.
With our observation distribution, the lane position can be accurately and efficiently estimated
using PF.
The proposed method is compared with Kalman filter and shows superior performance when
single detection result is not satisfactory.

The rest of the paper is organized as follows:
the problem of lane tracking is formulated in section \ref{sect:sys},
and a new observation model is introduced in section \ref{sect:obs}.
Section \ref{sect:PF} discusses how to track with the new model using
particle filter.
We present the experimental results in section \ref{sect:exp}
and conclude the article in section \ref{sect:conc}.

\section{System Model}
\label{sect:sys}

The goal of our proposed system is to detect two immediate lane markings on both sides (left and right) of the vehicle. Lane markings are usually painted in sharp contrast to the surrounding colors.

For the sake of simplicity, each lane marking is modeled as a straight line. In a curved lane it means the tangent line on the curve in near view. This approximation works well with more gradual varying curves. A line is specified by $(\rho, \theta)$, where $\theta$ is the angle made between the line and x-axis and $\rho$ is its distance to the origin (upper left corner).
The state vector $X_t$ here is composed of ${\rho}$, ${\theta}$,
as well as their first order derivative $v_{\rho}$ and $v_{\theta}$:
\begin{equation}
	X_t=\left[\rho_t\ v_{\rho_t}\ \theta_t\ v_{\theta_t}\right]
\end{equation}

To estimate the lane state sequentially, we need a dynamic model to describe its temporal evolution.
Constant-velocity model ($X_{t+1} = X_t + v_tT$) is the simplest motion model
that produces decent results assuming the vehicle is moving slow, frame rate is high and the demand for accuracy is not too high.

State equation:
\begin{equation}
\label{equ:dynamic}
	X_{t+1}=F_tX_t+U_t
\end{equation}
where
\begin{equation}
	F_t=\begin{bmatrix}
			1 & T & 0 & 0\\
			0 & 1 & 0 & 0\\
			0 & 0 & 1 & T\\
			0 & 0 & 0 & 1
		\end{bmatrix}
\end{equation}
and
$F_t$ is the state transition matrix linking the state vector at time step $t$ to time step $t-1$.
$T$ is the time gap between consecutive frames, the value being 0.06s.
$U_t$ is the process noise conforming to Gaussian process $\mathcal{N}(0,Q_t)$.
With simple derivation \cite{SINGER70}, the process noise covariance matrix $Q_t$ for the constant velocity model is given by:
\begin{equation}
	Q_t=\begin{bmatrix}
	\begin{bmatrix}
	T^3/3 & T^2/2\\
	T^2/2 & T
	\end{bmatrix}\cdot\sigma_{\rho_t}^2 &
	\begin{bmatrix}
	0 & 0\\
	0 & 0
	\end{bmatrix} \\
	
	\begin{bmatrix}
	0 & 0\\
	0 & 0
	\end{bmatrix} &
	\begin{bmatrix}
	T^3/3 & T^2/2\\
	T^2/2 & T
	\end{bmatrix}\cdot\sigma_{\theta_t}^2
	\end{bmatrix}
\end{equation}
where $\sigma_{\rho_x}$ and $\sigma_{\theta_x}$ are standard deviation of acceleration. Their values can be calculated by first taking the second order derivative of ground truth data and then calculating the standard deviation.

\section{Observation}
\label{sect:obs}

\subsection{Lane Detection}

The goal of detection stage is to extract useful lane marking features $(\rho, \theta)$ from the raw image ($1280\times720$). The first step is transforming the image from RGB space into HSV space, because HSV information best represents the perception through human vision system \cite{SPAIN}. Two relatively distant points in RGB space can appear very similar as seen by a human and vice versa. HSV space resolved this issue. We also shrink the image size down to ($640\times368$) to save computation time.

The next step is applying an averaging filter to smooth out the imperfections in the raw image. The imaging system of the video camera introduces some salt and pepper noise which must be removed before applying gradient detection. Otherwise too many edge points are detected. The averaging filter $\left[ \begin{array}{ccccc} \frac{1}{5} & \frac{1}{5} & \frac{1}{5} & \frac{1}{5} & \frac{1}{5} \end{array} \right]$ is chosen small enough to avoid blurring the gradient transition on real edges.

\begin{figure}
  \centering
    \begin{tabular}{cc}
        \includegraphics[width=4.5cm,height=3cm]{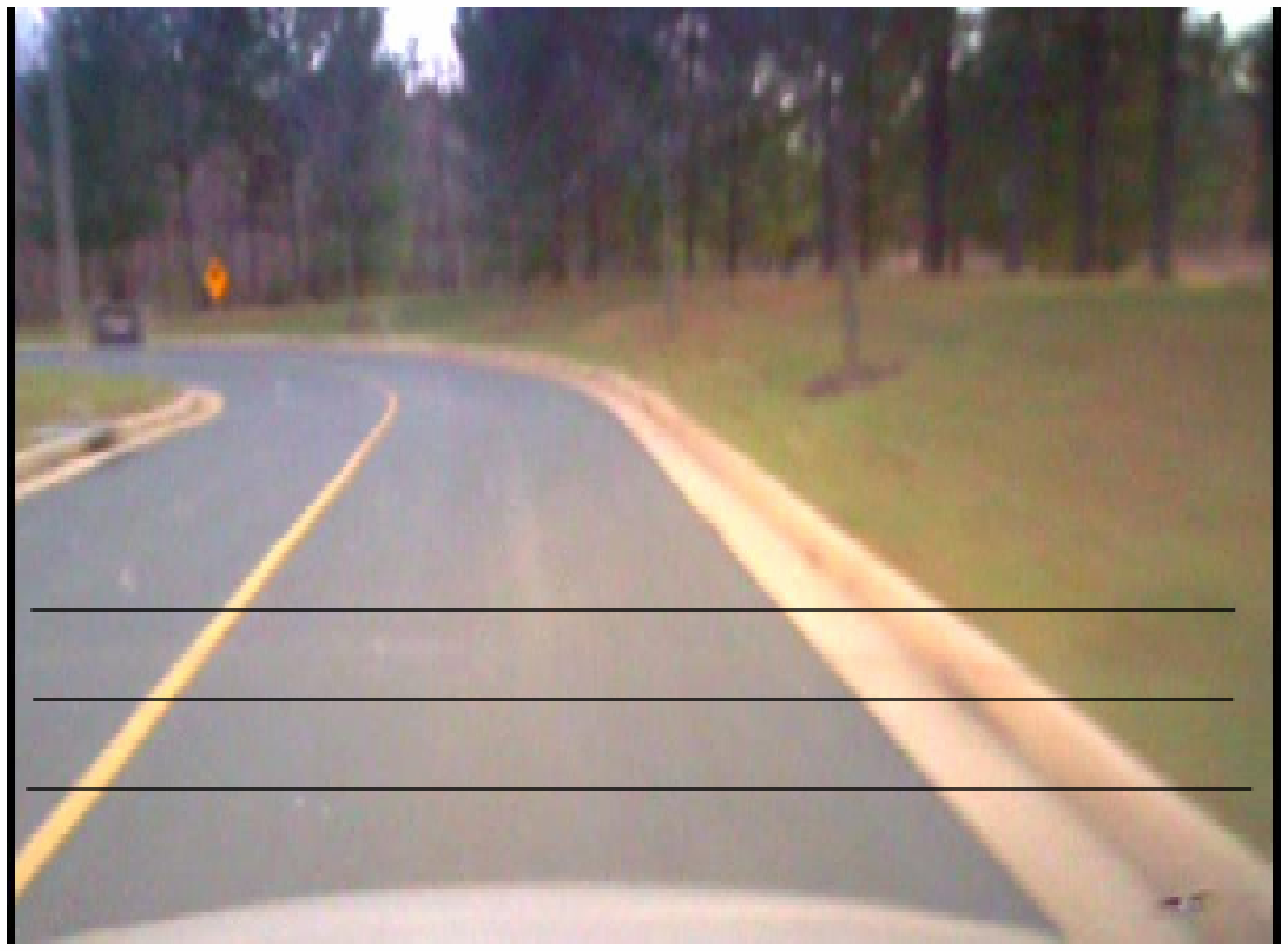} &
        \includegraphics[width=6cm]{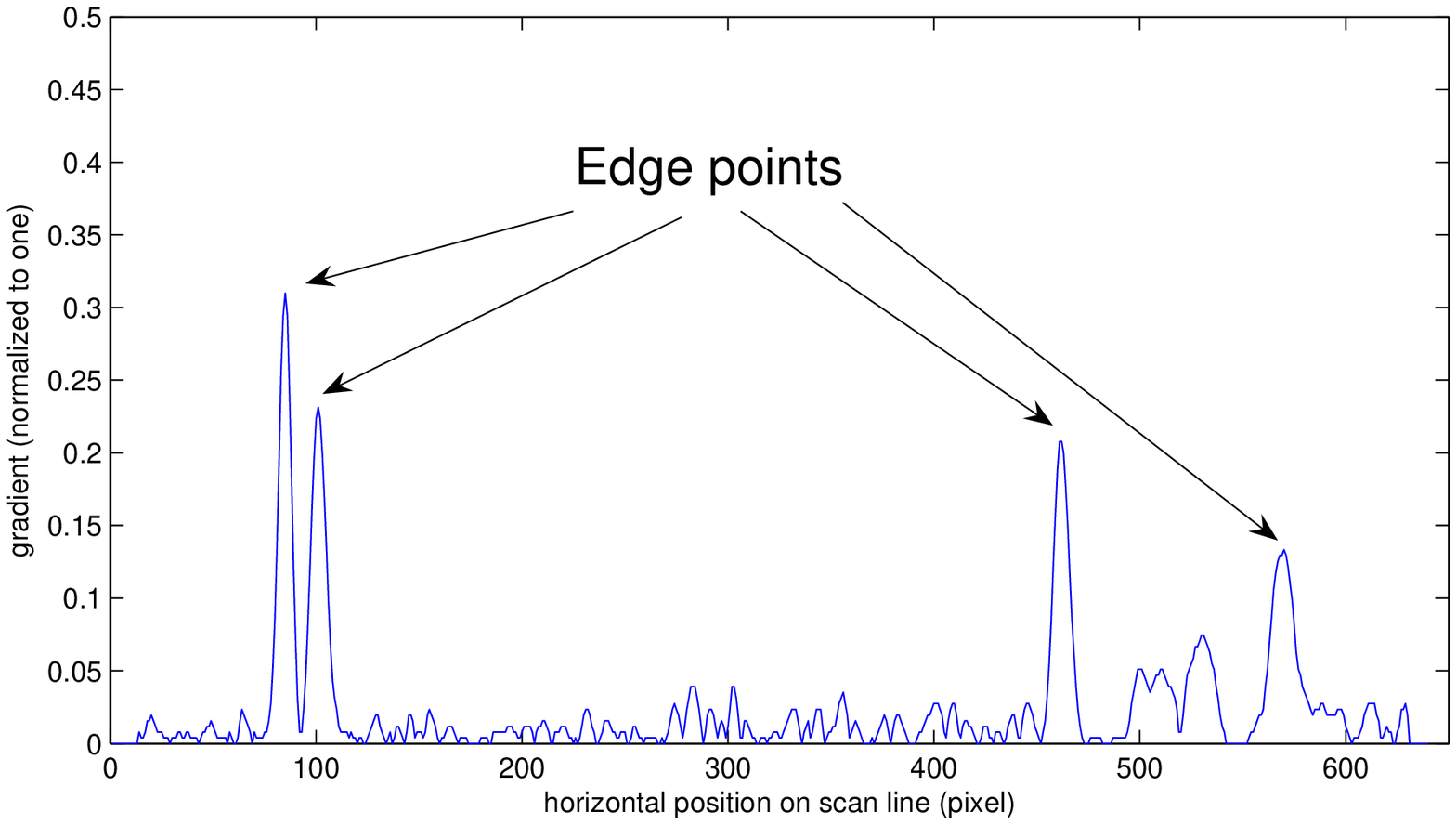} \\
        (a) & (b)
    \end{tabular}
    \caption{Edge detection (a) horizontal scan lines on a raw image;
             (b) output of gradient detector on the bottom scan line in (a).}
    \label{fig:detect}
\end{figure}

\begin{figure}
\centering
\includegraphics[width=8cm]{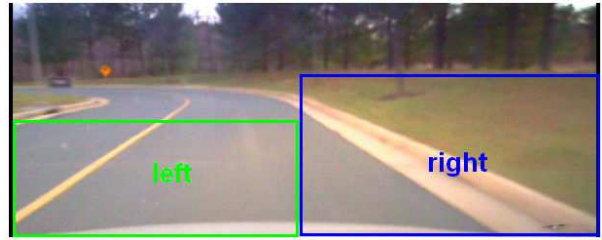}
\caption{Region of interest}
\label{fig:ROI}
\end{figure}

The gradient detector uses horizontal scan lines (Fig. \ref{fig:detect}) to examine the changes in pixels' $H$, $S$ and $V$ values and to segment the image into edge points and non-edge points. The stencil for gradient calculation is $\left[ \begin{array}{ccccccc} -1 & 0 & 0 & 0 & 0 & 0 & 1 \end{array} \right]$. Combined with previous averaging filter, the stencil produces an equivalent gradient calculator of $\left[ \begin{array}{ccccccccccc} -\frac{1}{5} & -\frac{1}{5} & -\frac{1}{5} & -\frac{1}{5} & -\frac{1}{5} & 0 & \frac{1}{5} & \frac{1}{5} & \frac{1}{5} & \frac{1}{5} & \frac{1}{5} \end{array} \right]$, i.e., the difference between the average of five pixels to its right and the average of five pixels to its left. The scan line is then swept across the Region of Interest (Fig.~\ref{fig:ROI}) to generate all edge points.

Pixels with gradient above a certain threshold level are marked as edge points. The output after this step is a 0-1 binary image.

Region of Interest (ROI) (Fig.~\ref{fig:ROI}) localizes search space to where observations are expected, so that unnecessary computation can be reduced. Following ROI, we use classic Hough Transform to find lines from the binary image. The detected lines are subject to $\rho$ and $\theta$ limiting. For example, lines with $\theta$ too small or too large are not considered. The line with the highest accumulator value, i.e., the most visible line is chosen as the single observation for KF.

\subsection{Gaussian Observation Model}

Observation equation:
\begin{equation}
	Y_t=H_tX_t+V_t
\end{equation}
\begin{equation}
	H_t=\begin{bmatrix}
						1 & 0 & 0 & 0\\
						0 & 0 & 1 & 0
			\end{bmatrix}
\end{equation}
where $H_t$ is the observation matrix, which observes only $\rho_t$ and $\theta_t$ but not their derivatives which is difficult without a digital speedometer.
$V_t$ denotes the observation noise, which is a Gaussian process $N(0,R_t)$.

Left markings and right markings use separate observation models and are assumed to be uncorrelated.

Experiments show that $R_t$ is not always a fixed value. Instead, $R_t$ varies between $\begin{bmatrix}1 & 0\\0 & 1\end{bmatrix}$ and $\begin{bmatrix}10 & 0\\0 & 10 \end{bmatrix}$, because the environment is dynamically changing (e.g., when the car has just entered a curve). Thus the noise cannot be modeled using a fixed Gaussian distribution. Even if the distribution is Gaussian, it is extremely hard to accurately estimate its covariance due to lack of ground truth in many cases.

If only one observation is chosen from the observation model, the succeeding filtering process will inevitably face a big challenge. It is usually possible for a specific scene to find a detection algorithm that produces observations whose noise has close to zero mean and very small variance. But this technique normally hurts generality. For a detection algorithm to work reasonably well in vastly different scenes, it has to give up scene-specific information. The resulting observations are likely to be poor and non-Gaussian.

To demonstrate that observation noise can be non-Gaussian under a generic detection algorithm, we conducted an experiment using three videos. Experimentally it is difficult to draw multiple samples from the distribution of $V_t$ at a fixed time, but sampling the same process over time is relatively easy to achieve. Fortunately for stationary ergodic processes, the mean and variance do not change over time, and time average of a conforming process equals the ensemble average. Thus it is reasonable to use the samples drawn from different times to represent the distribution at any instance in time.

\begin{figure}
  \centering
    \begin{tabular}{cc}
        \includegraphics[height=4.5cm]{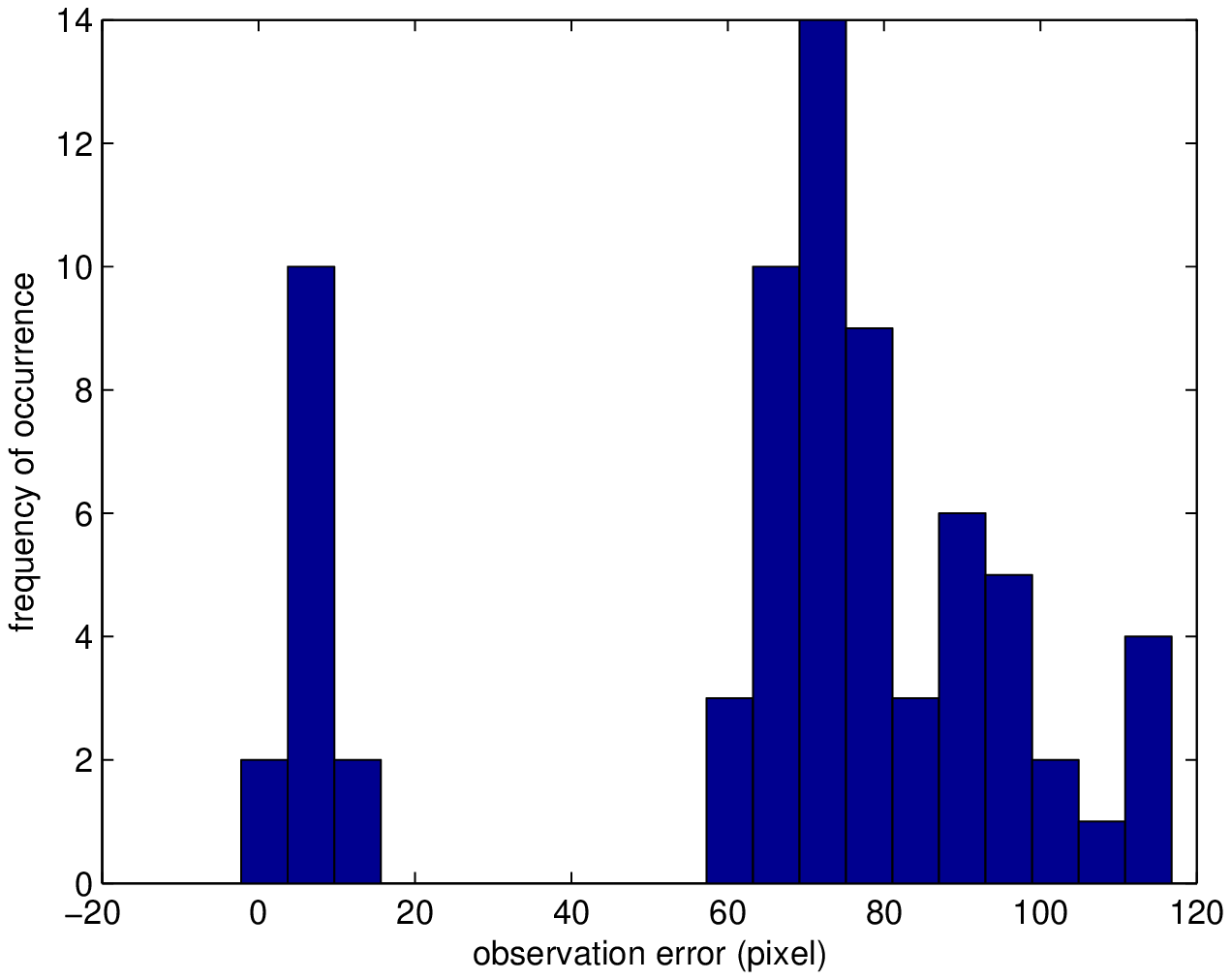} &
        \includegraphics[height=4.5cm]{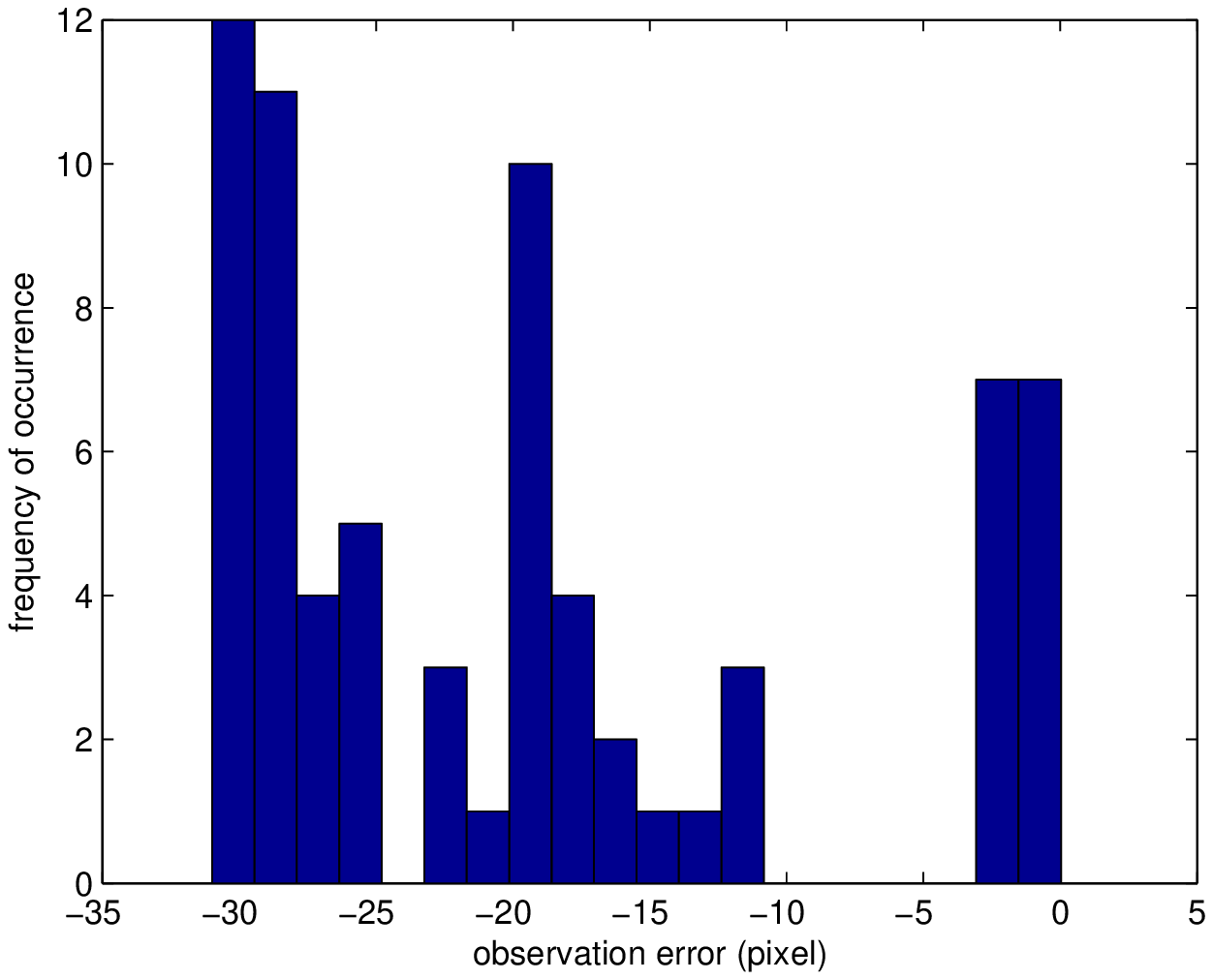} \\
      (a) & (b) \\
    \end{tabular}
    \caption{$V_t$ distribution (a) video C, $\rho_{right}$; (b) video C, $\theta_{right}$}
    \label{fig:VtDistri}
\end{figure}

These resultant distribution lacks the symmetry of a Gaussian distribution.
The main reason why observation noise becomes non-Gaussian is that only a single observation is selected. Noisy road condition usually has many false observations that overshadow the true ones. These false observations typically contain little information about the actual state (while true observations are random samples of a Gaussian distribution centered around the truth location). Averaging multiple observation is also prone to failure because the average will fall between the truth and the false edges, making an improbable observation.

\subsection{Multiple Mode Observation Model}

Since multiple observations can improve the quality of tracking,
we will set a threshold for the accumulator value: any line scoring higher
than the threshold becomes an observation mode.
By applying a relatively low threshold in the lane detection stage, we can
obtain a set of $K$ candidate lane parameters as our observation
$Y_t = \{ y_t^k, w_t^k \}_{k=1...K}$, where $y_t^k=(\rho_t^k, \theta_t^k)$ is the $k$'s
detected lane at time $t$, and $w_t^k$ is the associated weight to be defined
in the following.
We propose to fuse the multiple detection results in one unified observation
distribution, so that it can more authentically capture the image information
and at the same time has a compact form.
Gaussian Mixture Model (GMM) is a good choice in this regard, so we define
the new multi-mode observation distribution as a $K$'s order GMM:
\begin{equation}
\label{equ:observation}
    p(Y_t | X_t) = \sum\limits_{k=1}^K w_t^k
            \mathcal{N}\left([\rho, \theta]^T; y_t^k, \Sigma \right)
\end{equation}
where $\mathcal{N}$ denotes a normal distribution, and $\Sigma$ is a predefined
covariance matrix.

The weight $w_t^k$ denotes the probability of a detected lane being truth, which
favors certain observations over others.
This is advantageous in the presence of false edges.
While false edges cannot be eliminated during the Hough Transform stage,
we can formalize a function to penalize them according to their $\rho$ and $\theta$.
In this work, the weight is expressed as:
\begin{equation}
	w_t^k \propto \frac{1}{d_{car} \cdot d_{focus}}
\end{equation}
where $d_{car}$ specifies the distance from the vehicle to the line. It can be estimated as the line's intercept on the bottom edge of the image. The closer the intercept is to the midpoint, the smaller distance. Lines further away from the vehicle are assigned smaller weights as they are not imminent information. While looking down on the road from above, all lane markings are parallel. These parallel lines in the image coordinates all intersect at one fixed focus point in the image. That focus point remains its location relatively stable from frame to frame, unless the vehicle forms a very big angle with the lane. $d_{focus}$ is the distance from the focus point to the line. Lines with increased distance from the focus point are not parallel to normal lane markings

\section{Kalman Filter}

Kalman Filter (KF) has been used extensively in tracking a moving object with a known dynamical model.
In the aforementioned dynamic and observation model, if we assume $U_t$ and $V_t$ are drawn from Gaussian distribution and $F_t$ and $H_t$ are known linear functions, then KF gives the optimal estimate in terms of minimum mean square error.
KF works in two-step process. In the prediction step, it uses the dynamic model Eq. (\ref{equ:dynamic}) to obtain the prior pdf of the state at time $t+1$ ($\hat{X}_{i+1 | i}$). A following update step uses the observation at $t+1$ ($Y_{t+1}$) to compute the posterior pdf according to Bayes' rule.

Many applications have underlying state space model that is nonlinear Gaussian. Extended Kalman Filter (EKF) \cite{PFTutor} and Unscented Kalman Filter (UKF) \cite{UKF} have been proposed to solve these systems. Since the lane tracking problem in our case has a linear model choosing the basic KF will suffice. However, none of these variants of KF are capable of dealing with non-Gaussian noise present in our problem.

\section{Particle Filter}
\label{sect:PF}

To deal with the non-Gaussian observation distribution in Eq. (\ref{equ:observation}),
we propose to use Particle Filter (PF) for state tracking,
which is general enough to approximate posterior density function in any non-Gaussian
non-linear systems.

Particle filtering \cite{PFTutor} is a technique to find the discrete approximation of state
posterior within dynamic Bayesian framework.
Suppose we want to estimate a sequence
of hidden state variable $X_{0:t}$ based on observations $Y_{1:t}$.
As the observation model here is non-Gaussian, there will be no
analytic expression for the posterior, and we have to approximate it
with discrete samples (or particles) in the Monte Carlo way:
\begin{equation}
\label{equ:MonteCarlo}
    p(X_{0:t}|Y_{1:t}) \approx \sum\limits_{i=1}^{N_s}
                            w_t^i \delta(X_{0:t}-X_{0:t}^i)
\end{equation}
where $\{X_{0:t}^i\}$ is the set of $N_s$ randomly sampled
particle sequences, and $\{w_t^i\}$ is the associated weights.
Particles drawn from an importance density
$q(X_{0:t} | Y_{1:t})$ can be factorized as:
\begin{equation}
    q(X_{0:t} | Y_{1:t}) = q(X_{0:t-1} | Y_{1:t-1})
        \times q(X_t | X_{0:t-1}, Y_{1:t})
\end{equation}
then, the weights can be updated recursively as:
\begin{equation}
\label{equ:WeiUpdate1}
    w_{t}^{(i)} \propto w_{t-1}^i \frac{p(Y_t | X_t^i) p(X_{t}^i | X_{0:t-1}^i)}
                {q(X_t^i | X_{0:t-1}^i, Y_{1:t})}
\end{equation}
When the importance density is set to be the same as the dynamic model,
which is the case for Sequential Importance Resampling (SIR) \cite{SIRPF},
Eq. (\ref{equ:WeiUpdate1}) can be simplified as:
\begin{equation}
\label{equ:WeiUpdate2}
    w_t^i \propto w_{t-1}^i p(Y_t | X_t^i)
\end{equation}

Now given the process model in Eq. (\ref{equ:dynamic}) and the observation model
in Eq. (\ref{equ:observation}), we can readily track the lane state $X$ with
the particle filter procedures as summarized in Table \ref{tab:PF}.

\begin{table}
\centering
\caption{Algorithm of Lane Tracking by Particle Filter}
\label{tab:PF}
\begin{tabular}{p{100mm}}\hline
\vspace{-5mm}
\begin{itemize} \setlength{\itemsep}{-\parsep}
    \item for ($i =1:N_s$)
    \begin{itemize}
        \item Propagate particle $X_{t-1}^i$ according to Eq. (\ref{equ:dynamic}),
              and get $\tilde{X}_t^i$
        \item Evaluate the observation likelihood $p(Y_t | \tilde{X}_t^i)$
              by Eq. (\ref{equ:observation})
        \item Update weight $w_t^i$ by Eq. (\ref{equ:WeiUpdate2})
    \end{itemize}
    \item endfor
    \item Sort $\{ \tilde{X}_t^i, w_t^i \}_{i=1...N_s}$ according to weight $w_t^i$
    \item Resample (if necessary) $\{ \tilde{X}_t^i, w_t^i \}_{i=1...N_s}$,
          producing un-weighted sample set $\{ X_t^i, 1/N_s \}_{i=1...N_s}$
\end{itemize} \nointerlineskip \\ \hline
\end{tabular}
\end{table}

\section{Implementations and Results}
\label{sect:exp}

\subsection{Experiment Setup}
Images are captured at a constant rate of 16 frames per second from a low-cost digital camera mounted on the rear-view mirror inside the vehicle. All the calculations in this work are done in the image coordinates.

Five video sequences (named A through E), each with a different scene, are used as inputs to our program. Some of them have distracting objects by the roadside that may create false detections. The lengths of these sequences range from 49 frames to 80 frames.

We manually annotated the images with true lane markings from driver experience. Then ground truth states ($\rho_{truth}, \theta_{truth}$) are extracted from these images. ($\rho_{truth}, \theta_{truth}$) are compared with filter output from KF and PF, and mean square error (MSE) is computed as a metric to measure tracking quality.

We have to define the behavior of our filters in case of missing observations. For KF, only the prediction step is performed as it depends only on the previous estimate and the dynamic model.
While for the observation model Eq. (\ref{equ:observation}) used in PF,
missing observation is a special case with $K=0$ lanes detected.
And the multiple mode observation reduces to uniform distribution.

Most of the algorithm are implemented as C++ source code using OpenCV library. A small part of KF is implemented in Matlab. Following is a screenshot of the tracking result at run time.

\subsection{Results Analysis}

\begin{figure}
  \centering
  \includegraphics[height=5cm]{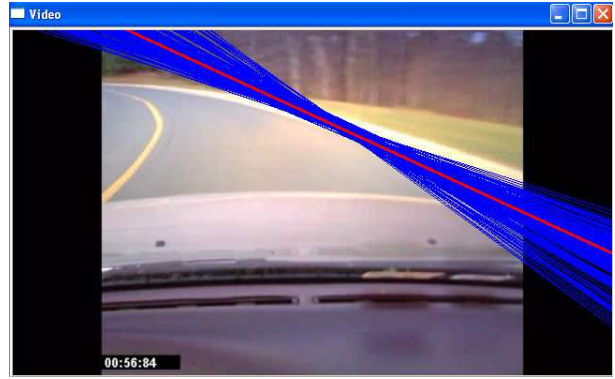}
  \caption{Results of right lane tracking with particle filter.
            All particles are visualized as blue lines, and their mean value is plot in red.}
  \label{fig:OpenCV}
\end{figure}

\begin{figure}
  \centering
    \begin{tabular}{cc}
        \includegraphics[height=4.5cm]{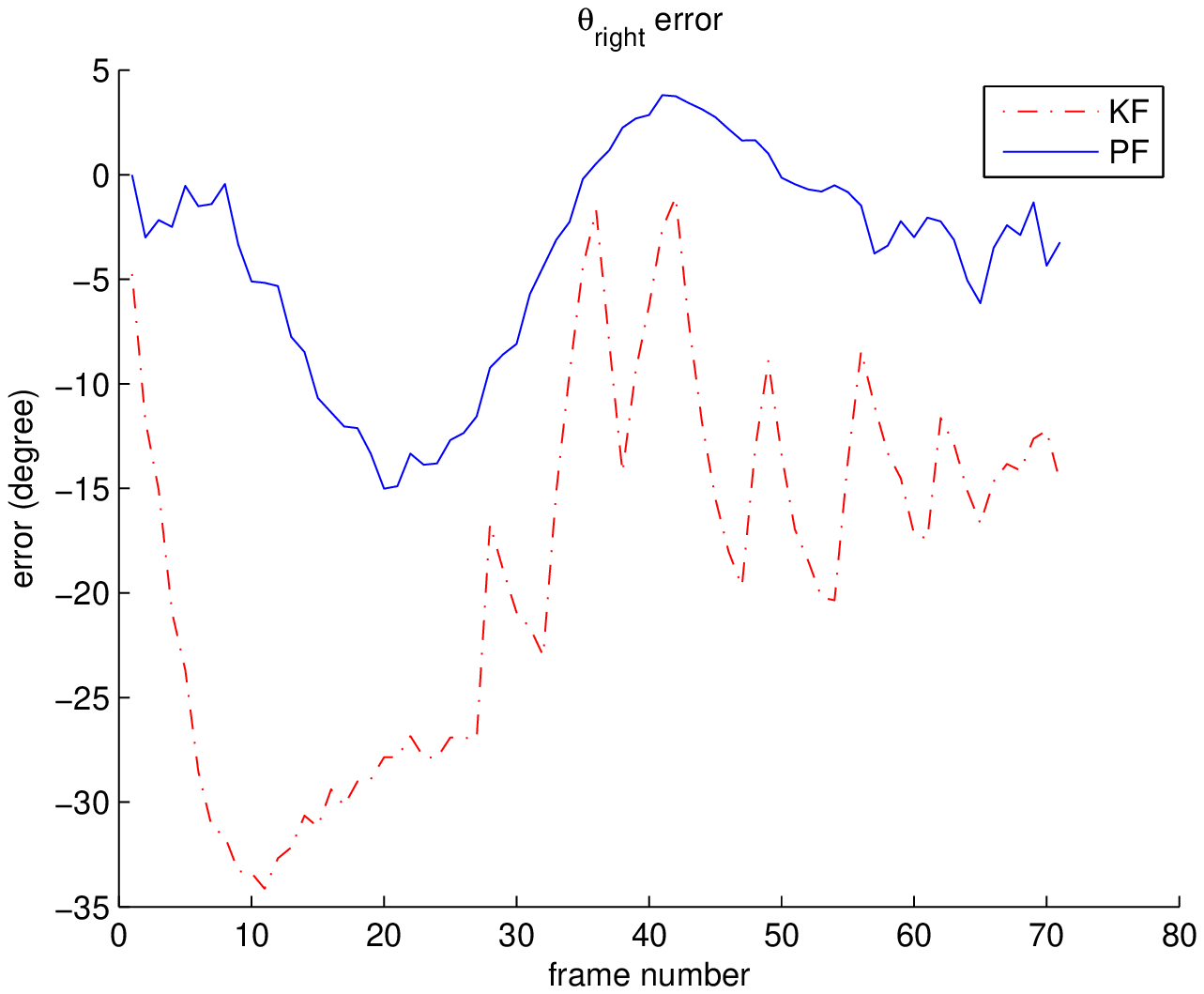} &
        \includegraphics[height=4.5cm]{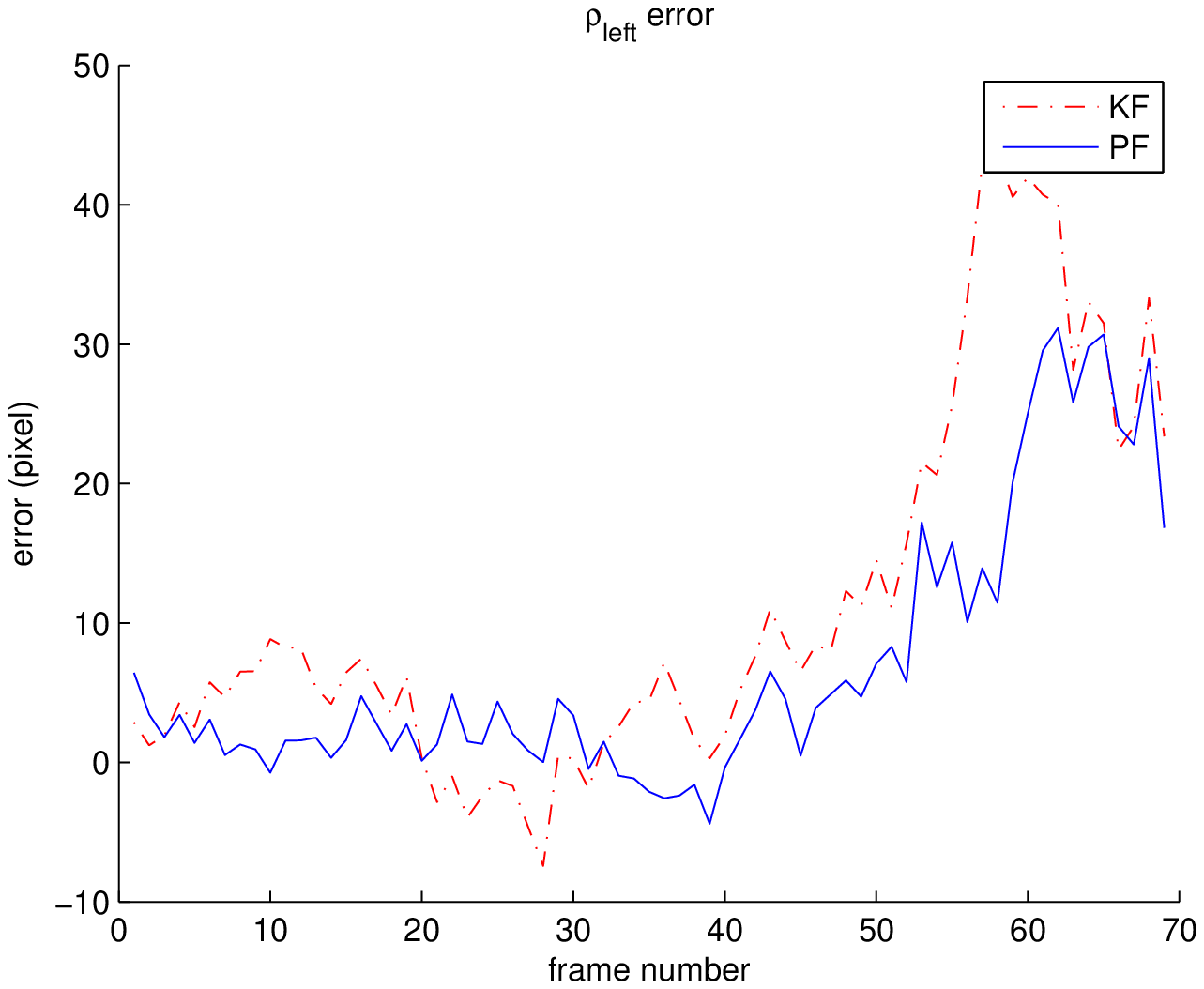} \\
      (a) & (b) \\
    \end{tabular}
    \caption{Instant tracking errors of KF and PF on different state variables:
             (a) $\theta_{right}$ in video C; (b) $\rho_{left}$ in video D}
    \label{fig:OnlineErr}
\end{figure}

Fig. \ref{fig:OpenCV} shows a typical output of our PF lane tracker (in red line),
which matches the truth lane quite well.
The instant tracking errors against ground truth for some state variables are plotted in
Fig. \ref{fig:OnlineErr}. Compared with KF, PF generally features a smaller error.

Table \ref{tab:MSE} lists the mean square error for both KF and PF on all the video sequences.
As we can see from the table, PF has slightly higher mean error than KF in several results, but
it usually happens when the overall error is small and does not represent any problem (the rho and theta resolutions are 1 pixels and 1 degrees respectively, so two errors are indistinguishable if they are within resolution apart). KF performs well only when the process and observation noise follow a Gaussian-like distribution.
Nevertheless, PF outperforms KF in most other cases as KF starts to deteriorate.

It should be pointed out that,
in video A, the main road is connected with several side roads on both sides. Lane markings are missing near the intersections. As a result, the detection algorithm is likely to take a strong false edge from background objects as observations.
If these false observation persist for multiple frames, KF will converge to those observations and is unable to recover.
Similar situations exist in video B and C.
In such cases, the multiple model observation is more reliable and the output of PF tracker
is more robust to false detection.

\begin{table}
\centering
\caption{Tracking Mean Square Errors of KF and PF}
\label{tab:MSE}
\begin{tabular*}{0.55\textwidth}
{@{\extracolsep{10mm}}c||cr@{.\extracolsep{0mm}}l@{\extracolsep{10mm}}r@{.\extracolsep{0mm}}l}
\hline
Video & State & \multicolumn{2}{c}{KF} & \multicolumn{2}{c}{PF} \\
\hline
\raisebox{-5.0ex}[0cm][0cm]{A}&
  $\rho_{left}$ & 28&58 & 21&43 \\
& $\theta_{left}$ & 2&25 & 3&35 \\
& $\rho_{right}$ & 46&52 & 27&11 \\
& $\theta_{right}$ & 13&34 & 9&24 \\
\hline
\raisebox{-5.0ex}[0cm][0cm]{B}&
  $\rho_{left}$ & 9&73 & 9&29 \\
& $\theta_{left}$ & 1&76 & 2&68 \\
& $\rho_{right}$ & 50&49 & 14&0 \\
& $\theta_{right}$ & 10&42 & 6&69 \\
\hline
\raisebox{-5.0ex}[0cm][0cm]{C}&
  $\rho_{left}$ & 3&3 & 3&79 \\
& $\theta_{left}$ & 2&15 & 3&48 \\
& $\rho_{right}$ & 72&25 & 19&68 \\
& $\theta_{right}$ & 20&62 & 6&48 \\
\hline
\raisebox{-5.0ex}[0cm][0cm]{D}&
  $\rho_{left}$ & 17&38 & 11&62 \\
& $\theta_{left}$ & 3&22 & 2&18 \\
& $\rho_{right}$ & 35&84 & 5&91 \\
& $\theta_{right}$ & 9&39 & 2&36 \\
\hline
\raisebox{-5.0ex}[0cm][0cm]{E}&
  $\rho_{left}$ & 11&0 & 9&12 \\
& $\theta_{left}$ & 3&22 & 1&73 \\
& $\rho_{right}$ & 2&41 & 3&25 \\
& $\theta_{right}$ & 0&79 & 1&38 \\
\hline
\end{tabular*}
\end{table}


\section{Conclusion}
\label{sect:conc}

The problem of video lane tracking is considered in this article.
A novel multiple mode observation model is put forth by incorporating multiple
detection hypothesises in a GMM distribution, so that more information can
be learned from source image.
The non-Gaussian observation is dealt with particle filter in sequential tracking.
Experimental results on various real videos firm that the proposed observation
model can promote the tracking accuracy notably.


\end{document}